\documentclass[letterpaper]{article}
\usepackage{iccc}

\usepackage{graphicx}

\usepackage{times}
\usepackage{helvet}
\usepackage{courier}
\title{The societal and ethical relevance of computational creativity}
\author{Michele Loi\\
University of Zurich   \\
  \texttt{michele.loi@ibme.uzh.ch} \\\And
  Eleonora Vigan\'{o}\\
  University of Zurich  \\
  \texttt{eleonora.vigano@ibme.uzh.ch} \\\AND
  Lonneke van der Plas \\
  University of Malta \\
  \texttt{lonneke.vanderplas@um.edu.mt}
}

\begin{document} 
\maketitle
\begin{abstract}
In this paper, we provide a philosophical account of the value of creative systems for individuals and society. We characterize creativity in very broad philosophical terms, encompassing natural, existential, and social creative processes, such as natural evolution and entrepreneurship, and explain why creativity understood in this way is instrumental for advancing human well-being in the long term. We then explain why current mainstream AI tends to be anti-creative, which means that there are moral costs of employing this type of AI in human endeavors, although computational systems that involve creativity are on the rise. In conclusion, there is an argument for ethics to be more hospitable to creativity-enabling AI, which can also be in a trade-off with other values promoted in AI ethics, such as its explainability and accuracy.    
\end{abstract}

\section{Introduction}
Creativity is beneficial to many aspects of human life, from the natural to the individual, to the social level. Current mainstream AI systems risk hindering processes that enable creativity and thus risk to reduce their benefits. In this contribution, we argue for the ethical and social benefits of creativity, identify the potential threats of AI for creativity, and present AI systems that enable creativity. The scope of our argument concerns both creative AI (i.e. AI generating valuable novelty directly) and creativity-enabling AI (i.e. AI that enables or favors the generation of valuable novelty by humans). We reason that by creating a bridge between the fields of AI ethics/philosophy and computational creativity (CC), and highlighting the threats of mainstream AI on human well-being, the societal uptake of CC can be promoted and lead to focal research in the field.

Most philosophical definitions of creativity involve the creation of something novel \cite{gaut_philosophy_2010} i.e. originality. Something can be new in the sense of it being the first time that it has been produced in history (H-creativity) or in a person’s life (P-creativity) \cite{boden_creative_1991}. The dominant traditions see creativity as essentially related to the production of something valuable \cite{gaut_philosophy_2010}, i.e. creativity is essentially not just novelty, but valuable novelty. We shall not assume that a process condition - e.g. a special form of independence from past models \cite{kronfeldner_creativity_2009} - defines what creativity is. On the contrary, we shall show that certain processes just happen to favor creativity, i.e. favor or enable the production of valuable novelty. Thus the attribution of creativity to certain processes is based on observations and does not result from a choice of definition.

\section{Processes enabling creativity}
\textbf{Natural creativity.} The  natural evolution of living entities fits our definition of a creative process.  This view has been influential for attempts to model creativity computationally \cite{campbell_blind_1960} as evolution, clearly, involves the production of novelty  (H creativity) \cite{boden_creativity_2018},
which is valuable for the organism (i.e. adaptive) or as a means to humankind. This kind of creativity is Darwinian, in that it is based on blind variation and subsequent selection. “Blind” genetic variation may not be “random”, but it is at least “un-directed”\cite{boden_creativity_2018}.

\noindent\textbf{Social creativity}. The collective dimension of creativity is enabled by liberty, both social and economic liberty. Social liberty allows for individual and collective “experiments of life” \cite{mill_liberty_1859} and for discovering new forms of social value previously thought impossible, e.g. stable relationship grounded in homosexual love and sexual conduct. The different forms of the good, in Mill’s view, cannot be discovered by pure intellectual acts of understanding, but must be lived out concretely \cite{anderson_john_1991}. 
Creativity is also expressed by markets (the creative destruction of capital) \cite{schumpeter_capitalism_1965}. Markets resemble natural systems, even if product innovations are not random or undirected, but on the whole they are rather bets on future success. Since most entrepreneurs do not know whether their enterprise will succeed, the market is (in the short term) a highly inefficient system: predicting company success is highly inaccurate  
 and 90\% of startups fail. Yet, in order to produce the novelties that we  highly value, ”[n]atural  and nature like systems want some overconfidence on the part of individual  economic  agents, [...]  provided that their failure does not impact others  [...]”\cite{taleb_antifragile:_2012}.  Individual entrepreneurial ignorance works like a high-risk bet, which enables exploration (i.e. of unknown and unpredictable consumer preferences) and is ultimately
 responsible for generating valuable novelty for consumers.

\noindent\textbf{Individual existential creativity.} In the individual dimension, creativity requires the individual’s attitude of exploring life possibilities and experimenting life plans and versions of herself. In this process, the individual deepens the knowledge of herself, acquires life experience, and tests the life track that she previously chose. Individual existential creativity is thus not exclusively the purview of the artist, but of any individual that adopts the exploration of life opportunities and her potential as a pivotal value in her life. Individual existential creativity refers to how the individual is leading her life; therefore, it can be assessed with the standards of the philosophical conception of prudence. While, in the common meaning, prudence is conceived as a cautious attitude toward risk and danger, in philosophy, it is the pursuit of one’s own  good throughout a lifetime. Individual existential creativity contributes to “philosophical prudence” as it is a way to live one’s life and pursue one’s good. One fundamental attitude that enables creativity is the openness to the unbidden, which is a disposition of recognition and acceptance that not everything is, or should be, under one's control \cite{sandel_case_2007}. The openness to the unbidden favors the acceptance of a world sometimes characterized by extreme variability and a mental disposition in which not everything has to be planned in detail. For example, the inclination of a parent to accept his or her child, however unexpected and different the child ends up being from the parent, accommodates biological variance. Creativity in life also involves the individual's experimentation with different possible selves, namely trying various life tracks, each realizing a different idea of herself. Typically, in our culture, the experimentation phase has been confined to adolescence and the consolidation and exploitation phase to all later life stages. However, it is likely that in the future, with an increasing acceleration of social change and technology, life-plans will be built in such a way that experimentation phases can also occur later in life. This will require society to both encourage openness and experimentation during people's lifespans and governments to provide a safety net to help them start again, in case of failure.

\begin{figure}[!ht]
\begin{center}
\includegraphics[scale=0.35]{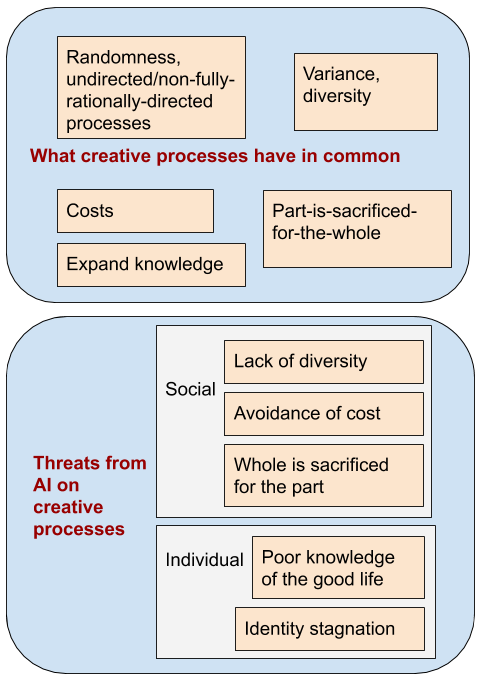} 
\caption{Creative processes and threats}
\label{pic-overview}
\end{center}
\end{figure}

\subsection{What these processes have in common}
Summarizing the cross-domain analysis so far, we can list the following features of processes enabling creativity:
\begin{enumerate}
 \item some degree of randomness, or at least, undirected processes, or at the very least non-fully-rationally-directed processes (e.g. processes directed in ways that are rationally defective or that lack rationality altogether). 
At the natural level, we have random mutations; at the social level, we have free initiative in markets and creative pursuits; and at the individual level, we have experimentation with a set of values and commitments that do not have a guaranteed payoff. 
 \item the generation of more variance than in the case of non-random, directed, and fully rational processes. 
For example, new life forms emerge through mutations, new needs are discovered when new products and services are invented, and new forms of life and values are discovered through social experimentation.
 \item the generation of costs, that is, additional costs compared to non-random, directed, and fully rational processes. These processes are wasteful, and can even generate harm (e.g. harmful mutation).
At the natural level, life evolves in spite (and because of) many failed experiments. At the social level, resources are wasted, for example by failed businesses. At the individual level, the costs of creativity lay in the inefficient pursuit of a plan, commitment, career, etc. In fact, when the individual pursues a high-risk plan, she may end up with less satisfaction, happiness, and resources than with a low risk-plan, and the costs of recovery may be high. 
 \item  limited reach of the damage, and “part-is-sacrificed-for-the-whole” principle. 
Most mutations are harmful but biological processes are flexible, most start-ups fail but consumers benefit, some experiments of life may be harmful to individuals, but societies learn that new forms of the human good are possible. At the individual level, while some life experiments pursued in a specific phase of the individual’s life may fail, subsequent experiments would typically benefit from self-knowledge and knowledge of reality gained with  earlier unsuccessful trials. 
\end{enumerate}

\noindent Of these four features, (3) is positively harmful, (4) is beneficial, (1) and (2) are valuable as means. Our hypothesis is that both non-directedness and variance are means to exploring the space of possibilities (and generating knowledge), in particular, possible biological forms, preferences and needs, and shapes of the good human life. At the individual level, what gets explored is the individual's potential and how to realize it. This is crucial for self-knowledge and authenticity, which are arguably essential to well-being \cite{sumner_welfare_1996}. In conclusion, creativity is the production of valuable novelty, which contributes to generating knowledge. The latter is especially valuable for humans as it enables them to adapt to the rather unpredictable world outside them and discover who they, collectively and individually, are.

\section{The threat of current mainstream AI systems } 
Collective creativity is threatened by AI technologies that currently prevail. Let us take economic liberty as an example. We explained how the individual's ignorance regarding the utility function and the willingness to take risks are important for the system to create novelty. A society in which decision making is highly informed by intelligent systems that are trained in a supervised fashion with a narrowly defined utility function that tries to avoid risk will not exhibit the same exploration power as a system based on the individuals' judgments. Moreover, in this new, efficiency-driven reality, fewer agents will be taking over the decision making that was previously done by many more individuals. This leads to uniformity in the decision making process, which will lead to less diversity in the collective “experiments of life” we referred to earlier. However, these are crucial for creative processes in society. Lastly, we see an additional threat in the training cycle of these systems. They learn from past data labelled with relevant outcomes. The more intelligent systems come to influence decision making in society, the more they will impoverish their own training data, as the future data that serves to train them will over time become increasingly compatible with the original system expectations, since the data itself is influenced by these systems. The data will be more and more uniform, leaving many possible regions of the search space unexplored. 

At the individual level, many decisions affecting the individual’s 
opportunities and plans are implemented by systems based on AI such as recommendation systems and digital wellness technologies. These systems influence the individual’s choice in various ways, by means of the visual presentation of the alternative as well as by the alternatives that are suggested by the system. These systems have two features that result in identity stagnation. First, current AI technologies maximize the present self’s utility function and thus present the individual with choice alternatives that satisfy her current preferences. This implies that such technologies make identity and preference changes less likely. Second, the profiling of the future self provided by current AI systems is based on past data \cite{Mittelstadt2016}. The latter excludes unlikely but disruptive events from the predictions of the algorithms; therefore, current AI systems nudge the user to continue on the current life track. This 
identity stagnation 
decreases the individual’s overall well-being by limiting the possibility to experiment with new life tracks and learn from them, and make the attitude of openness to the unbidden useless, as the user is never provided with unexpected options.

\section{Computational systems involving creativity}

However, not all work in AI is harmful for creative processes, on the contrary, and the ICCC conference series are the living proof of that. Although for this short paper, an exhaustive survey is out of scope, we would like to give a short (and certainly incomplete) overview of computational systems that involve aspects of creativity. To organize this wide spectrum of systems, we can determine what aspect of creativity they are focusing on, the creative product or the creative process, which are two main aspects of study (Said-Metwaly \shortcite{Said2017Methodological}, among others). 
Computational work that focuses on the first aspect includes work on the generation of unconventional linguistic variations, such as the generation of metaphors \cite{veale-hao-2008-fluid}, scientific ideas \cite{wang-etal-2019-paperrobot}, or visual art \cite{Elgammal2017CANCA}. Although automatically generating creative output requires putting thought into the creative process, the main focus of these works is on the creative output.

We would like to zoom in on computational work, where the focus is on the creative process. Some work builds computational models to better understand the cognitive process of creativity. 
Works from the field of cognitive science, for example, have shown evidence that creative people have more complex semantic network structure and may activate a wider range of associations across the network than less creative people \cite{Kenettetal2014}. 

Other work focuses on creating computational algorithms that introduce aspects of novelty in the learning process. Many are from the field of robotics, for example, they describe situations in which robots need to navigate in an unknown environment and need to look for novelty in order to better explore the space and not get stuck in local optima. One type of solutions solves this task by defining a so-called intrinsic reward (based on the psychological concept of curiosity (Barto \shortcite{barto2013intrinsically}, among others).  Intrinsically Motivated Reinforcement Learning (Kaplan and Oudeyer \shortcite{KaplanOudeyer2006} among others) basically works as follows: the agent is rewarded for discovering new patterns in the environment \cite{Schmidhuber2010}, and it is always in search of novelty.  Other work aims at modelling social creativity specifically \cite{Saunders2019MultiagentbasedMO}. The author reports on experiments, in which multiple agents with diverse hedonic functions, which determine levels of interest and actions to be taken, work together. 
Another strand of research introduce aspects of novelty in the learning process under the header of evolutionary computing (EC). An approach known as Novelty Search \cite{lehman2011abandoning} searches for behavioral novelty instead of seeking an objective. 
Also here, we see EC algorithms that are built to run in a distributed fashion over a population of agents and works in which evolution occurs within one agent only, usually employing a time-sharing strategy of genes.
 
In summary, we find several algorithms that involve aspects of creativity. These algorithms exhibit the four characteristics of creative processes we listed previously: aspects of undirectedness, generation of variance, generation of costs in order to increase knowledge, and the part-is-sacrificed-for-the-whole principle. They also implement these both at the individual level of single agents and at the level of social interaction. However, these algorithms are not the mainstream algorithms that find their way to the market, where narrowly defined objective-driven high-accuracy systems prevail. Although even here we recently see some changes taking place. Recommender systems, such as video recommendation systems for Youtube, incorporate reinforcement learning, and off-policy learning to avoid 'myopic recommendations', where the short term reward overshadows long-term user utility in the form of discovery content \cite{Ma2020}. Motivations here come from trying to optimize the long-term user utility. We feel that attention to the ethical aspects, related to the threats the current AI technologies pose, could spearhead the work on computational creativity. 

\section{Conclusion}

Individual and collective creativity are ethically valuable because they are essential to (a) adaptation to the unexpected; (b) self-knowledge. Both adaptation and self-knowledge are "permanent interests of man as a progressive being" \cite{mill_liberty_1859}. Adaptation is a permanent interest of humanity because humans face environments that are unpredictable in the long term. Self-knowledge is essential to well-being because a life cannot be good \textit{for} an individual unless it reflects her individuality, and individual preferences must be informed (including, by self-knowledge) in order for happiness to be authentic \cite{sumner_welfare_1996}.

Current AI ethics guideline documents \cite{jobin_global_2019} mention freedom and autonomy as higher-order principles. But in those guidelines the focus is on protecting \textit{human} freedom and autonomy, typically against the overreach of poorly controlled AIs. When freedom and autonomy are addressed, there is no reflection on the idea that the autonomy and creativity of AIs may also be needed as enablers of human freedom, autonomy, and well-being. In our analysis, computational creativity may be needed for human well-being and it is thus in a trade off with other legitimate goals of AI (e.g. accuracy). We hope that this brief reflection motivates scholars of computational creativity to reflect on the ethical importance of their discipline and contribute to AI ethics by researching open questions such as, how to build systems that integrate computational creativity with resilience and how to avoid excessive harm.

\section{Acknowledgments}
This work was partly funded by the Digital Society Initiative of the University of Zurich and the Cogito Foundation grant 17-117-S. We would like to thank the reviewers, and Paul O. Dehaye and Albert Gatt for their valuable suggestions.






\bibliographystyle{iccc}
\bibliography{iccc}

\end{document}